\documentclass[letterpaper, 10 pt, conference]{ieeeconf} 

\IEEEoverridecommandlockouts            
\usepackage{cite}
\usepackage{amsmath}
\usepackage{amssymb}
\usepackage[pdftex]{graphicx}
\usepackage[dvipsnames]{xcolor}
\usepackage[hyphens]{url}
\usepackage{caption}
\usepackage{algorithm}
\usepackage{algpseudocode}
\usepackage{layouts}
\usepackage{color}

\usepackage{tabularx}
\usepackage{hyperref}

\newcolumntype{P}[1]{>{\centering\arraybackslash}p{#1}}
\newcolumntype{M}[1]{>{\centering\arraybackslash}m{#1}}

\DeclareCaptionFont{mysize}{\fontsize{8}{9.6}\selectfont}
\title{\LARGE \bf
Enhancing Human-Robot Collaboration Transportation through Obstacle-Aware Vibrotactile Feedback}

\author{Doganay Sirintuna$^{* 1,2}$, Idil Ozdamar$^{* 1,2}$, Juan M. Gandarias$^{1}$, and Arash Ajoudani$^{1}$
\thanks{$^{*}$ These authors contributed equally to this work.}
\thanks{$^{1}$ Human-Robot Interfaces and Interaction Laboratory, Istituto Italiano di Tecnologia, Genoa, Italy. {\tt \{doganay.sirintuna, idil.ozdamar, juan.gandarias, arash.ajoudani\}@iit.it}}
\thanks{$^{2}$ Dept. of Informatics, Bioengineering, Robotics, and System Engineering. University of Genoa, Genoa, Italy.}
\thanks{This work was supported in part by the ERC-StG Ergo-Lean (Grant Agreement No.850932), in part by the European Union’s Horizon 2020 research and innovation programme SOPHIA (Grant Agreement No. 871237).}
}

\begin{document}
\maketitle
\thispagestyle{empty}
\pagestyle{empty}
\begin{abstract}
Transporting large and heavy objects can benefit from Human-Robot Collaboration (HRC), increasing the contribution of robots to our daily tasks and reducing the risk of injuries to the human operator. This approach usually posits the human collaborator as the leader, while the robot has the follower role. Hence, it is essential for the leader to be aware of the environmental situation. However, when transporting a large object, the operator's situational awareness can be compromised as the object may occlude different parts of the environment. This paper proposes a novel haptic-based environmental awareness module for a collaborative transportation framework that informs the human operator about surrounding obstacles. The robot uses two LIDARs to detect the obstacles in the surroundings. The warning module alerts the operator through a haptic belt with four vibrotactile devices that provide feedback about the location and proximity of the obstacles. By enhancing the operator's awareness of the surroundings, the proposed module improves the safety of the human-robot team in co-carrying scenarios by preventing collisions. Experiments with two non-expert subjects in two different situations are conducted. The results show that the human partner can successfully lead the co-transportation system in an unknown environment with hidden obstacles thanks to the haptic feedback. 
\end{abstract}

\section{Introduction}
\label{sec:introduction}

Collaborative robots have seen advancements to the point where they can safely operate with humans as a team without requiring physical barriers. These human-robot teams, which combine human cognitive skills with the precision and repeatability of the robots, are expected to be high-performance solutions in many sectors ranging from healthcare to manufacturing~\cite{ajoudani2018progress}.

To establish an intuitive collaboration similar to a human-human partnership, most of the existing approaches in the literature focus on understanding human intention and designing the robot to respond accordingly. Indeed, previous studies have demonstrated that effective teams that allow the robot to perceive the human operator could be employed to solve practical problems such as assembly~\cite{KRUGER2009628}, surface treatment~\cite{SOLANES2018528}, and sawing~\cite{Peternel}. 

\begin{figure}
	\centering
	\includegraphics[trim={0 0.3cm 0 0},clip]{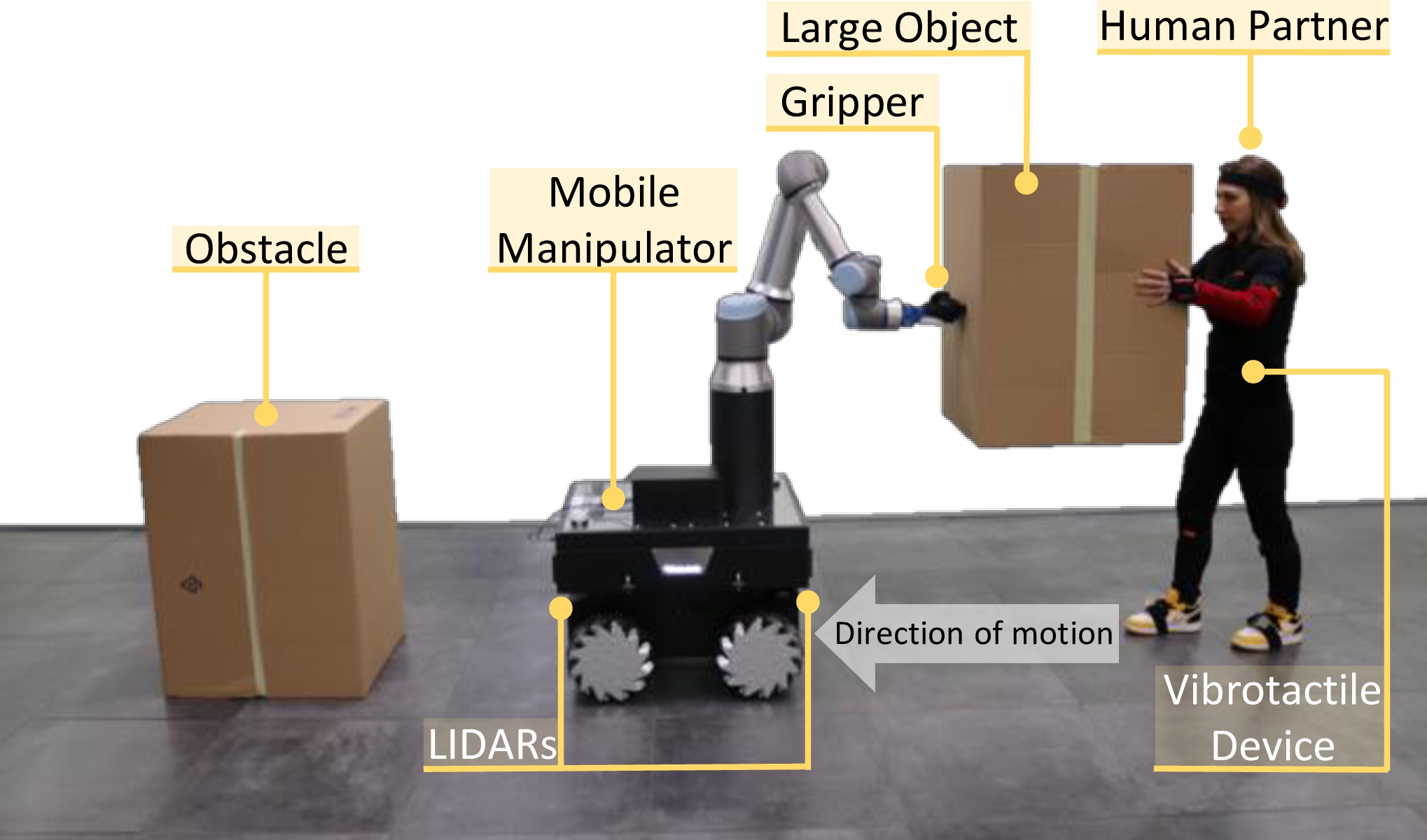}
	\caption{We propose a novel vibrotactile feedback-based awareness module that alerts the human operator about the obstacles in the environment during collaborative transportation.}
    \vspace{-0.7cm}
	\label{fig:cover}
\end{figure}

However, bi-directional communication between the partners can increase the human operator's awareness about the task's current state and the environment to increase safety and efficiency~\cite{Drury}. Different communication interfaces are employed in the literature depending on the work environment, the task, and the information type to be conveyed. For instance, in~\cite{doganaydrill}, an augmented reality (AR) interface was utilized to inform the human operator about the phases of the collaborative drilling task. 
The framework presented in~\cite{idil} enables controlling a configurable set of robotic arms thanks to a  graphical user interface (GUI) that displays information about currently active robots, the selected control strategy, and multiple options to reconfigure the robots according to the task needs.
As an alternative to visual interfaces, verbal feedback was provided in a collaborative task to increase team performance by augmenting the situation awareness of the human~\cite{stclair}. However, the drawback of this type of feedback is that it is not suitable in environments with high noise levels, such as manufacturing sites.

\begin{figure*}
	\centering
	\includegraphics[width=0.99\linewidth]{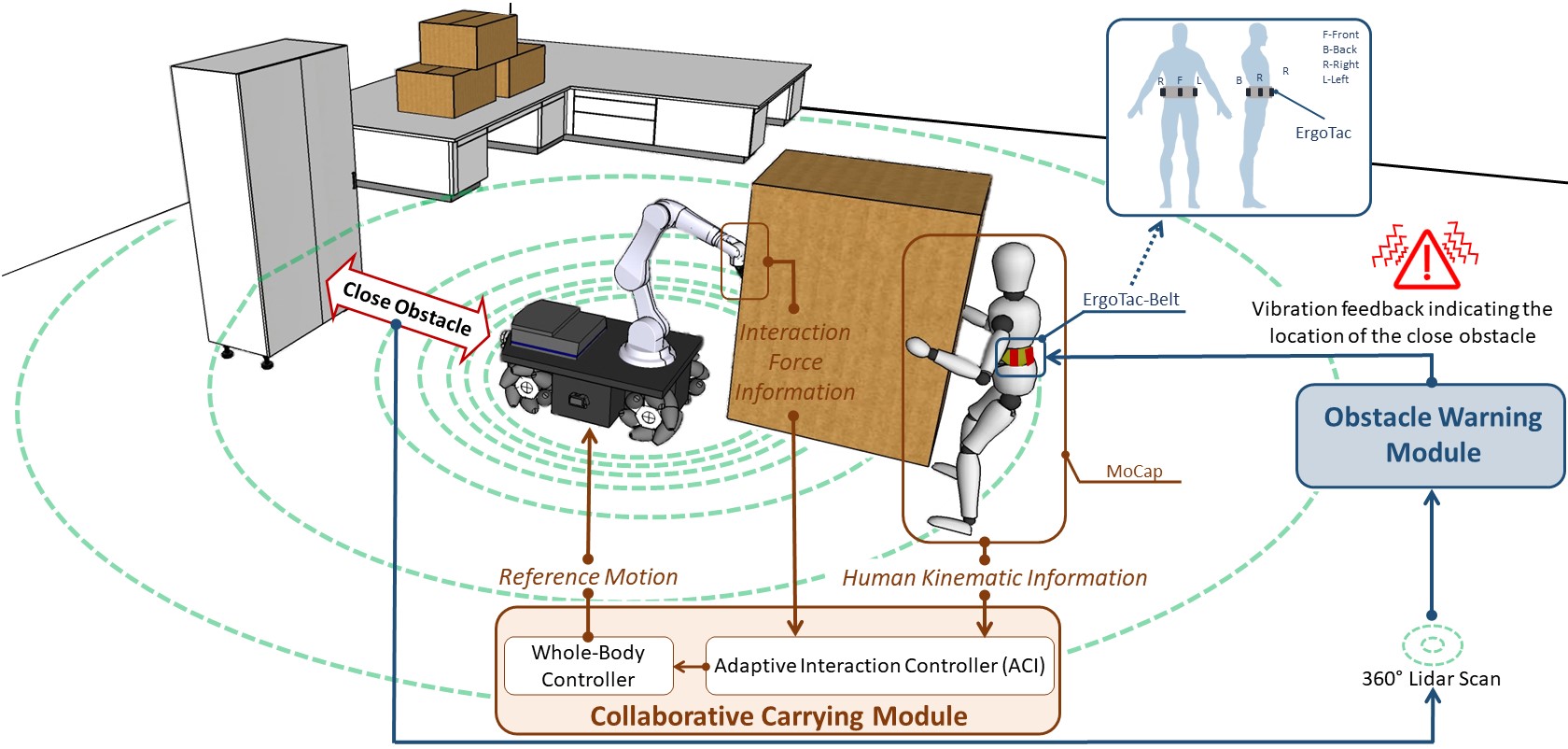}
	\caption{High-level scheme of the proposed framework. While the \emph{Collaborative Carrying Module} generates whole-body movements on a mobile collaborative robot by combining the haptic information transmitted through the object with the human kinematic information acquired from a motion capture system, the \emph{Obstacle Warning Module} produces vibratory feedback to the human through an ErgoTac-Belt indicating the detected obstacle's location and proximity by the LIDARs' scan.}
     \vspace{-0.7cm}
	\label{fig:system_overview}
\end{figure*}

Wearable vibrotactile devices are a promising alternative as they do not jeopardize the human senses, such as visual and auditory-based approaches~\cite{wearable_review}. In our previous work~\cite{ergotac}, we presented ErgoTac, a wearable device that uses vibration feedback to improve human ergonomics when performing heavy industrial tasks. Later, the ErgoTac-Belt, which employs 4 of the same devices, was utilized to guide the human center of pressure while walking~\cite{ergotac-belt}. Moreover, these kinds of tactile interfaces are also employed in the literature for human-robot teams. For example, Casalino et al. employed a vibrotactile ring to convey information to the user regarding the state of a collaborative assembly task~\cite{casalino}.

In this paper, we improve the collaborative transportation framework of our previous work~\cite{doganay2022} by introducing a novel awareness module that informs the human operator about the obstacles in the environment via ErgoTac-Belt (see Fig.~\ref{fig:cover}). In our previous work, a mobile manipulator and a human partner collaborated to carry out a transportation task purely led by a human. In this approach, occlusions when carrying large objects may occur, compromising the environmental awareness of the human and hence demoting safety. The new haptic-based obstacle-aware module alerts the human of any close obstacles that the robot's sensory system detects, including those behind the human or those that the transported item might occlude. Since the human operator is the pure leader in our co-transportation framework, augmenting the operator's awareness about the environment enhances the safety of the human-robot team by preventing potential undesired collisions.

\vspace{-0.2 cm}
\section{System Overview}
\label{sec:system_overview}

The interconnections between hardware and software components of our interactive co-transportation framework are shown in Fig.~\ref{fig:system_overview}. 
The employed robotic platform, which has a manipulator attached on top of its base, is driven by omnidirectional wheels that allow the platform to avoid non-holonomic constraints yielding movements over a large workspace.
It is equipped with an anthropomorphic robotic hand attached to the end-effector to co-carry objects with the human partner, a Force/Torque sensor at the flange to measure the wrenches resulting from the interaction, and LIDARs for inspecting the surroundings to detect obstacles.
The proposed framework consists of two main units: the \textit{Collaborative Carrying Module} and the \textit{Obstacle Warning Module}.   
The first one facilitates the transportation of objects with different deformation characteristics ranging from highly deformable to purely rigid thanks to the \textit{Adaptive Collaborative Interface (ACI)}, which we first presented in \cite{doganay2022} and then extended in \cite{doganayidil} to a heterogeneous human-multi-robot team. 
Inside this module, the ACI combines the interaction force information and the human kinematic information acquired from a motion capture system (MoCap) to generate reference control input of the robotic platform.
The second module generates vibrotactile feedback to the human via ErgoTac-Belt~\cite{ergotac-belt} according to the location and the proximity of the obstacles detected by the robot.
This way, the human partner's sensations are augmented, and they can be aware of the obstacles outside their field of view alongside the visible ones.

\vspace{-0.2cm}
\section{Methodology}
\label{sec:methodology}

\subsection{Collaborative Carrying Module}
\label{subsec:carrying_module}

\subsubsection{Whole-body Controller}
\label{subsubsec:whole_body_controller}
In this study, a weighted whole-body closed-loop inverse kinematics (CLIK) controller is utilized on the robot (see our previous work~\cite{doganay2022} for more details). This scheme calculates the desired joint velocities resulting in the desired end-effector motion while exploiting the redundancy of the robot. The primary and the secondary cost function are written as (dependencies are dropped): 
\begin{subequations}\label{eq:gfo}%
\begin{alignat}{2}%
&     \mathcal{L}_1 = || \dot{\boldsymbol{x}_d} + \boldsymbol{K}({\boldsymbol{x}_d} -{\boldsymbol{x}}) - \boldsymbol{J(q)}\dot{\boldsymbol{q}}||^2_{\boldsymbol{W}_{1}} + ||k\dot{\boldsymbol{q}}||^2_{\boldsymbol{W}_{2}}, \\& 
\mathcal{L}_2 = ||\boldsymbol{q}_{def} - \boldsymbol{q}||^2_{\boldsymbol{W}_{3}},
\label{eq:secondary_task}%
\end{alignat}%
\end{subequations}%
\noindent where $\boldsymbol{\dot{q}}$ $\in$ $\mathbb{R}^{n_{a}+n_{b}}$ is the optimization variable, $n_a$ and $n_b$ are the DoF of the base and the arm, $\boldsymbol{q}$ $\in$ $\mathbb{R}^{n_{a}+n_{b}}$ is the current joint positions, ${\boldsymbol{J}}$ $\in$ $\mathbb{R}^{6\times(n_{a}+n_{b})}$ is the whole-body geometric Jacobian, ${\boldsymbol{x}}$ $\in$ $\mathbb{R}^{6}$ is the current end-effector pose, ${\boldsymbol{W}_1}$ $\in$ $\mathbb{R}^{6\times6}$, ${\boldsymbol{W}_2}$, ${\boldsymbol{W}_3}$ $\in$ $\mathbb{R}^{(n_{a}+n_{b})\times(n_{a}+n_{b})}$, and ${\boldsymbol{K}}$ $\in$ $\mathbb{R}^{6\times6}$ are diagonal positive definite matrices, $k$ $\in$ $\mathbb{R}_{>0}$ is the so-called damping factor \cite{chiaverini1992weighted} and $\boldsymbol{q}_{def}$ is the default joint configuration. The calculated desired velocities by taking the negative gradient of the secondary task are projected onto the null-space of the first task to guarantee the desired-end effector motion while keeping the arm close to the base.

\subsubsection{Adaptive Collaborative Interface}

This interface is divided into two parts: the \emph{Admittance Controller}, which calculates a reference velocity based on the force transferred through the object, and the \emph{Reference Generator}, which sends the desired pose and twist to the Whole-body Controller by merging the velocity output of the Admittance Controller with the human hand velocity acquired from the MoCap system.

\paragraph{Admittance Controller}

We employ a standard admittance controller in the following form: 
\begin{align}
  \boldsymbol{F}_{ee} = \boldsymbol{M}_{adm}\dot{\boldsymbol{v}}_{adm}+ \boldsymbol{D}_{adm}\boldsymbol{v}_{adm},  
\end{align}
\noindent where $\boldsymbol{F}_{ee} \in \mathbb{R}^{3}$ is the measured forces at the end-effector, $\boldsymbol{M}_{adm}$ and $\boldsymbol{D}_{adm}$  $\in$ $\mathbb{R}^{3\times3}$ are the desired mass and damping matrices,   and $\boldsymbol{v}_{adm} \in \mathbb{R}^{3}$ is the admittance reference translational velocity.

\paragraph{Reference Generator}

During co-transportation, haptic information alone may not be enough for effective coordination between the human and the robot due to the deformability of the object. Our object-deformation agnostic controller handles this issue by combining the admittance reference velocity ($\boldsymbol{{v}}_{adm}$) and the human hand velocity ($\boldsymbol{{v}}_{h}$) based on an adaptive index ($\alpha$) to compute the desired velocity ($\boldsymbol{{v}}_{d}$), as follows:%
\begin{subequations}%
\begin{alignat}{1}%
&\alpha(t) =1-\frac{||\int_{t_{c}-W_{l}}^{t_{c}} \boldsymbol{{v}}_{adm}(t) \,dt||\ }{||\int_{t_{c}-W_{l}}^{t_{c}} \boldsymbol{{v}}_{h}(t) \,dt||\  + \epsilon} , \label{eq:alfa}\\& %
\boldsymbol{{v}}_{d}(t) = \boldsymbol{{v}}_{adm}(t) + \alpha(t)\boldsymbol{{v}}_{h}(t),
\end{alignat}%
\end{subequations}
where $\alpha \in [0,1]$ is the adaptive index, $t_c$ is the current time, $W_l$ is the length of the sliding time window and $\epsilon$ is a small number used to avoid the problem of division by zero. This adaptive formulation enables us to co-transport objects ranging from highly deformable ($\alpha(t) = 1$) to purely rigid ($\alpha(t) = 0$).

Next, the desired pose and twist are sent to the whole-body controller calculated by $\boldsymbol{{x}}_{d}(t) = \int_{0}^{t_c} \boldsymbol{\dot{x}}_{d}(t) \,dt $, where $\boldsymbol{\dot{x}}_{d}(t) = [{{v}_{d}(t)}^{T}, {0}^{T}]^T$.

\subsection{Obstacle Warning Module}
\label{subsec:warning_module}

\begin{figure}
	\centering
	\includegraphics[width=1\linewidth]{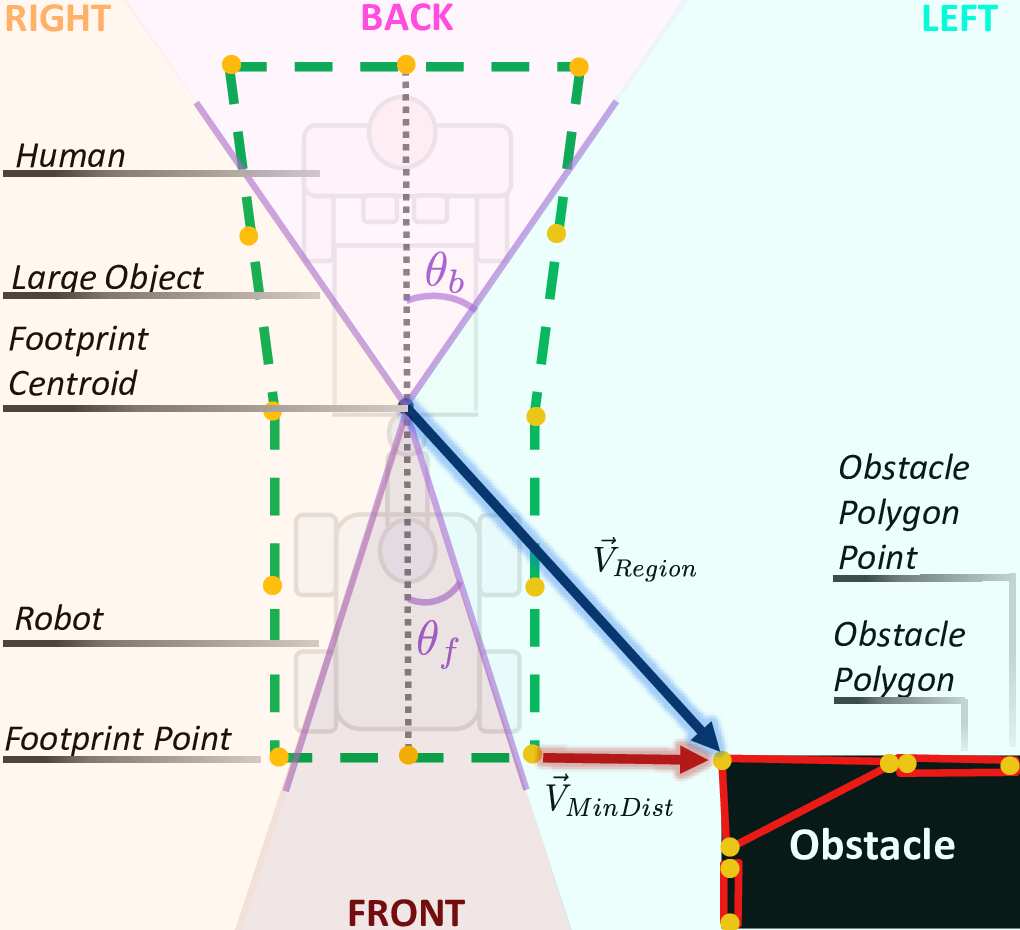}
	\caption{Illustration of the expanded footprint that encompasses both human and robot, the obstacle polygon, and the specified regions of the body relative directions.}
    \vspace{-0.7cm}
	\label{fig:footprint}
\end{figure}

This module aims to alert the human operator about the presence of obstacles in close proximity using the Ergotac-Belt, which consists of 4 vibrotactile feedback devices representing the body relative directions (see Fig.~\ref{fig:footprint}). To achieve this, firstly, this module detects the obstacles which might be occluded or behind the human using the equipped LIDAR sensors of the mobile base. Then, it generates appropriate vibrotactile feedback according to the detected close obstacles (if they exist).

In order to obtain information about the obstacles in the co-transportation environment, we employed the 'move\_base'\footnote{\url{http://wiki.ros.org/move_base}} package of the Robot Operating System (ROS). This package continuously updates a cost map by fusing data acquired from the perception sensors and odometry information without requiring an initial map. Then, a set of non-convex polygons is generated based on the occupied cells of this map~\cite{dbscan,concav}. This real-time updated approach allows us to obtain the positions of not only the static obstacles but also the dynamic ones. Note that, although this package can be employed to navigate the robot to a target location using the generated obstacle information, this functionality is not available for scenarios where the robot lacks the knowledge of the target, such as in our co-transportation task where the human is the leader.

During co-transportation, the human operator stands in front of the robot (see Fig.~\ref{fig:cover}), which can cause the algorithm to detect the human's legs as obstacles. To overcome this issue, we expanded the robot's footprint to encompass the human (see Fig.~\ref{fig:footprint}). Moreover, by doing so, the obstacles near the human are also taken into consideration, not only the ones close to the robot base. In fact, the distance to obstacles is calculated by determining the minimum distance pair between 12 selected points on the footprint and the generated obstacle polygon points (see $\vec{V}_{MinDist}$ in Fig.~\ref{fig:footprint}). However, we exploit the vector from the centroid of the footprint to the closest obstacle polygon point to find the region of the obstacle in terms of body relative directions (see $\vec{V}_{Region}$ in Fig.~\ref{fig:footprint}). Furthermore, we heuristically determined two separate angles for the front ($\theta_{f}$) and back ($\theta_{b}$) regions due to the difference in the width of the footprint between these two areas.

In this work, we decided to not vibrate more than one vibrotactile device at the same time as in our previous study~\cite{ergotac-belt}, in order to avoid the cognitive burden of the human operator. The pseudocode of selecting the appropriate ErgoTac device to be vibrated ($R_{t}$) and the intensity of the vibrations ($I_{t}$) according to the detected obstacles is given by Algorithm~\ref{alg:cap}. Here, we implemented a function to prevent fluctuations between different ErgoTacs when distances to obstacles in various regions are close to each other. This function enables switching to a new ErgoTac device if only the detected obstacle is closer than the one in the prior region by a specified '$Switch\_Ratio$'. To calculate the intensity of the vibrations, we utilize the following rules: \begin{equation}
I_t = \begin{cases} 
      1 & ||D_{{}_{w}O}||\leq d_{CRIT} \\
      1 - \frac{||D_{{}_{w}O}|| - d_{CRIT}}{d_{MAX} - d_{CRIT}} & d_{CRIT} < ||D_{{}_{w}O}|| \leq d_{MAX}. \\
     
   \end{cases}
 \label{Eqn:intensity}
\end{equation}
\noindent These rules vary the intensity of vibrations as a linear function of the distance between the closest pair of the obstacle polygon and footprint ($||D_{{}_{w}O}||$). The maximum intensity is achieved at a critical distance ($d_{CRIT}$) which is heuristically set to a low value, while the minimum occurs at the maximum distance ($d_{MAX}$), which serves as a threshold to filter the obstacles that are too far.

\renewcommand{\algorithmicrequire}{\textbf{Input:}}
\renewcommand{\algorithmicensure}{\textbf{Output:}}

\begin{algorithm}
\caption{Obstacle-Aware Vibrotactile Feedback}\label{alg:cap}
\begin{algorithmic}
\Require $R_{t-1}$
\Ensure $R_{t}, I_{t}$

\State $\boldsymbol{O} \gets getAllObstacles$
\State $\boldsymbol{O}_{th} \gets selectObstaclesUnderThreshold(\boldsymbol{O},d_{MAX})$

\If{$\boldsymbol{O}_{th}$ is empty}
    \State $R_{t} \gets None$
    \State $I_{t} \gets 0$
\Else
    \State $\hat{O} \gets findClosestObstacle(\boldsymbol{O}_{th})$
    \State $\boldsymbol{O}_{th}^{R_{t-1}} \gets findObstaclesInTheRegion(\boldsymbol{O}_{th},R_{t-1})$
    \If {$\boldsymbol{O}_{th}^{R_{t-1}}$ is empty}
        \State ${}_{w}O \gets \hat{O}$ \Comment{${}_{w}O$ : Obstacle to be warned}   
    \Else
        \State $\hat{O}^{R_{t-1}} \gets findClosestObstacle(\boldsymbol{O}_{th}^{R_{t-1}})$
        \If{$getObstacleRegion(\hat{O})$ equal to $R_{t-1}$}
            \State  ${}_{w}O \gets \hat{O}$
        \Else
            \State $||D_{\hat{O}}|| \gets getDistance(\hat{O})$
            \State $||D_{\hat{O}^{R_{t-1}}}|| \gets getDistance(\hat{O}^{R_{t-1}})$
            \If{$||D_{\hat{O}}||< ||D_{\hat{O}^{R_{t-1}}}||*Switch\_Ratio$}
                \State ${}_{w}O \gets \hat{O}$  \Comment{Switch}
            \Else
                \State ${}_{w}O \gets \hat{O}^{R_{t-1}}$ \Comment{No Switch}
            \EndIf
        \EndIf
    \EndIf
    \State $R_{t} \gets getObstacleRegion({}_{w}O)$
    \State $I_{t} \gets computeIntensity({}_{w}O)$ \Comment{Eq.~\ref{Eqn:intensity}}
\EndIf

\end{algorithmic}
\end{algorithm}
\vspace{-0.5 cm}

\section{EXPERIMENTS}
\label{sec:experiments}

\subsection{Experimental Setup}
\label{subsec:experimental_setup}

In order to validate the proposed framework, we used an experimental setup consisting of 3 main parts: (i) a robotic platform, (ii) a MoCap system, and (iii) a vibrotactile feedback device (see Fig.~\ref{fig:cover}). 
We employed the Kairos mobile manipulator as the robotic platform, which comprises an Omni-directional Robotnik SUMMIT-XL STEEL mobile base, a 6-DoFs Universal Robot UR16e arm with an F/T sensor to measure the applied wrenches at the robot's flange, a Pisa/IIT Softhand gripper, and two 2D SICK Microscan LIDAR sensors.
These sensors are located at the back and front of the robot base and 30 cm above the ground, providing a 360-degree sensing range. 
For the motion capture system (MoCap), the MVN Biomech suit (Xsens Technologies) is utilized since it provides accurate tracking of human kinematic information owing to its 17 inter-connected inertial measurement units (IMUs).
Note that our framework does not depend on specific hardware to access human kinematic information; hence alternative MoCap systems can be replaced with Xsens.
The ErgoTac-Belt is composed of 4 ErgoTac units located at the front (F), right (R), left (L), and back (B) of the operator at the L5 level (see Fig.~\ref{fig:system_overview}).
At the user side, robust and pleasant vibration feedbacks are generated thanks to ErgoTac's small dimension (68.1 mm × 37.0 mm × 17.3 mm), lightweight (28 g), and wireless communication feature with low energy consumption (multi-point connection via Bluetooth low energy at 2.4 GHz).

\subsection{Controller Parameters}
\label{subsec:controller_parameters}

The mass and damping of the \textit{Adaptive Collaborative Interface} were set to $\boldsymbol{M}_{adm}=diag\{12,12,12\}$ and $\boldsymbol{D}_{adm}=diag\{150,150,150\}$, respectively. The sliding time window length ($W_l$) used in Eq.~\eqref{eq:alfa} for calculating $\alpha$ was set to 0.5 seconds to accurately identify the object's deformation behavior causing a delay that could adversely influence the task performance.

In order to obtain precise tracking of the desired end-effector motion and avoid too high joint velocities, the $\boldsymbol{K}$, $\boldsymbol{W}_1$ and $\boldsymbol{W}_2$ parameters of the whole-body controller were chosen as $\boldsymbol{K}=diag\{0.1,0.1,0.1,0.01,0.01,0.01\}$, $\boldsymbol{W}_1=100\cdot diag\{10,10,10,5,5,5\}$ and $\boldsymbol{W}_2=diag\{\boldsymbol{5}_{n_b}, \boldsymbol{2}_{n_a}\}$  where $n_a=6$ and $n_b=3$.
Apart from that, to ensure locomotion behavior, most of the movement is assigned to the base instead of the arm by setting $\boldsymbol{W}_3$ to $diag\{\boldsymbol{0}_{n_b},\boldsymbol{3}_{n_a}\}$.

In this study, the front ($\theta_{f}$) and back ($\theta_{b}$) region angles were selected as 0.23 and 0.70 rad, respectively, by taking into account the shape of the footprint. The threshold distance ($d_{MAX}$) for filtering the obstacles located far from the robot was set to 1.1 m, while the critical distance ($d_{CRIT}$) that indicates the dangerously close obstacles was chosen as 0.2 m. Lastly, the $Switch\_Ratio$ was experimentally set to 0.8.

\begin{figure*}[t]
 \centering
    {\includegraphics[width=0.86\textwidth, trim=0cm 0cm 0cm 0cm, clip=true]{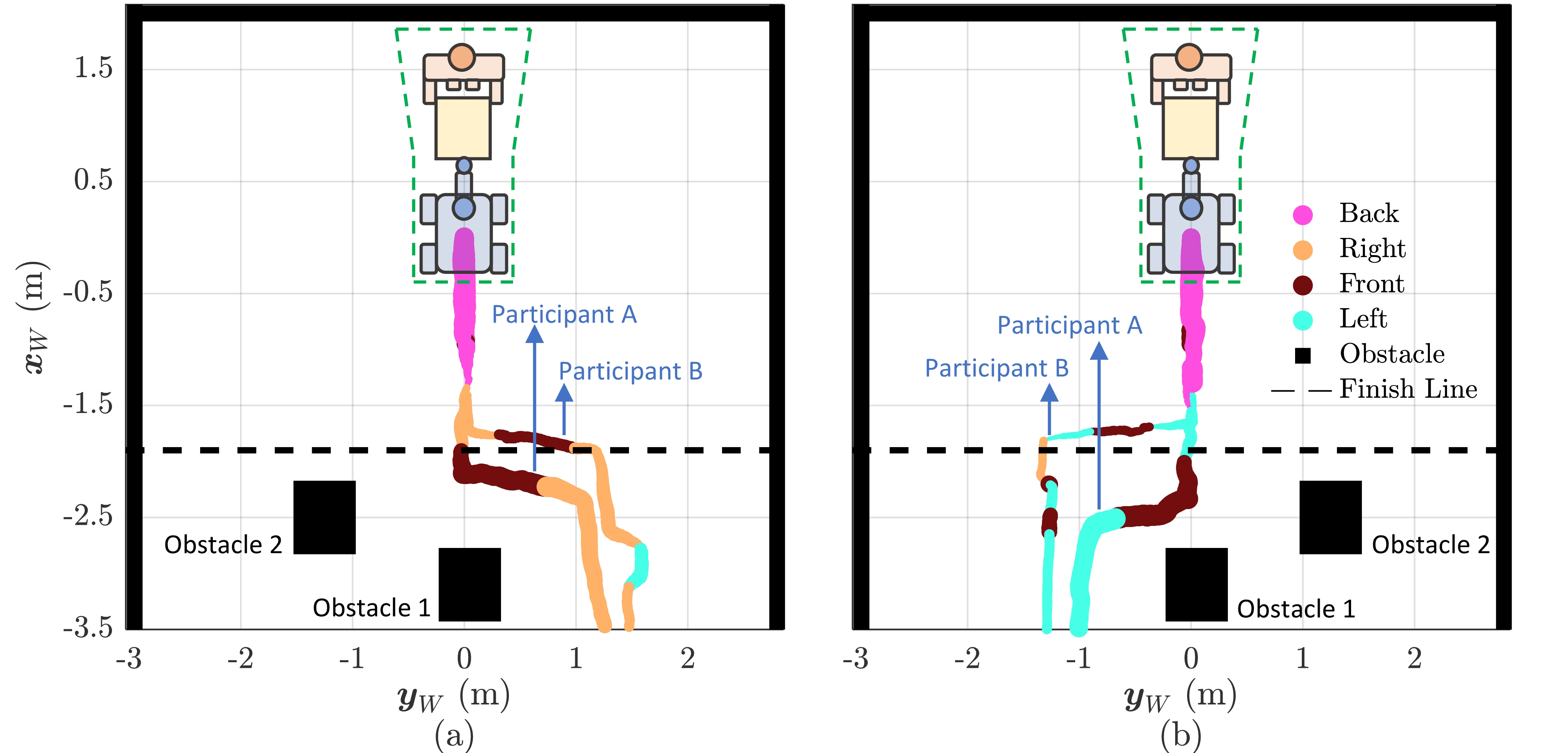}}
    \caption{The paths of the mobile base during the collaborative transportation task when obstacle 2 is on the right (a) and is on the left (b) w.r.t the human. While the colors indicate the region of the vibration feedback, the thickness shows its intensity.}
    \vspace{-0.7cm}
    \label{fig:experimens_with_path}
\end{figure*}

\subsection{Experimental Scenario}
\label{subsec:experimental_scenario}

The experiments were designed to evaluate the effectiveness of the \emph{Obstacle Warning Module}, as without the vibrations provided by it,  the operator could not navigate to the target location without having any collision. To this end, the participants were asked to co-carry an object until they reached the finish line that was located 3.4 m ahead of their initial point, where they were unaware of the positions of obstacles placed in the environment. The videos of these experiments with an additional demonstration where the operator is blindfolded are available at \url{https://youtu.be/UABeGPIIrHc}.

In the experiments,  a 1.9 kg box with dimensions of 80 × 60 × 50 cm was co-transported while the human and the robot were facing each other by holding the opposite sides of the box (see Fig.~\ref{fig:cover}).
The dimensions of the box were intentionally chosen such that it would block a substantial portion of the human's frontal and peripheral vision, making it challenging for them to see most of the surroundings while co-carrying the box.
During each trial, two obstacles (the exact dimensions of the box being carried) were placed in locations where they were not initially visible to the human and remained occluded for the majority of the experiment.
In the first task, obstacle 1 was positioned 4.3 m ahead of the participants, creating a scenario where the robot base would collide before they arrived at the target line (see Fig.~\ref{fig:experimens_with_path}a). 
Additionally, the second obstacle was positioned 3.7 m ahead and 1 m sidewise toward the operator's right. 
In the second scenario, the first obstacle remained in the same position; while the second one was located 3.7 m ahead and 1 m lateral to the left of the operator (see Fig.~\ref{fig:experimens_with_path}b).
Once the obstacles were placed, the participants were instructed to reach the finish line by taking the shortest path possible while maintaining a comfortable distance from the obstacles around.

The experiments were conducted with one female (26 years old) and one male (28 years old) volunteer in accordance with the Declaration of Helsinki, and the protocol was approved by the ethics committee Azienda Sanitaria Locale (ASL) Genovese N.3 (Protocol IIT\_HRII\_ERGOLEAN 156/2020).

\vspace{-0.2cm}
\section{RESULTS AND DISCUSSION}
\label{sec:results_and_discussion}

The dimensions of the room in which the experiment was carried out, the positions of the obstacles, the paths of the mobile base until the participants reached the finish line, and the feedback provided by the ErgoTac-Belt are shown in Fig.~\ref{fig:experimens_with_path}. 
The colors and thickness of the paths indicate the region (right, left, front, and back w.r.t. human), and the intensity of the vibration was transmitted through the belt.

At the beginning of the experiments, the participants felt vibration stimuli due to their close proximity to the wall (see Fig.~\ref{fig:experimens_with_path}a). During the forward motion of the human-robot dyad, the intensity of this vibration decreased. Then, the vibrated region altered to the right and began to increase as the robot approached obstacle 2 from the side. Here, participant A maintained the forward movement until the vibrated region changed to the front, and the intensity became significantly stronger, indicating that there exists a very close obstacle behind the robot. In contrast, participant B preferred to change the route in the same situation while the right side of the ErgoTac-Belt was still vibrating. Since the participants were aware that the right side was already occupied due to prior warnings, both of them chose to move left. After the vibration shifted to the right once again, both participants started their forward movement to arrive at the target line. Whereas participant A took a more direct route, participant B had a slight deviation to the left until ErgoTac-Belt generated feedback regarding the wall.

\begin{figure}[t!]
	\centering
	\includegraphics[width=0.99\linewidth]{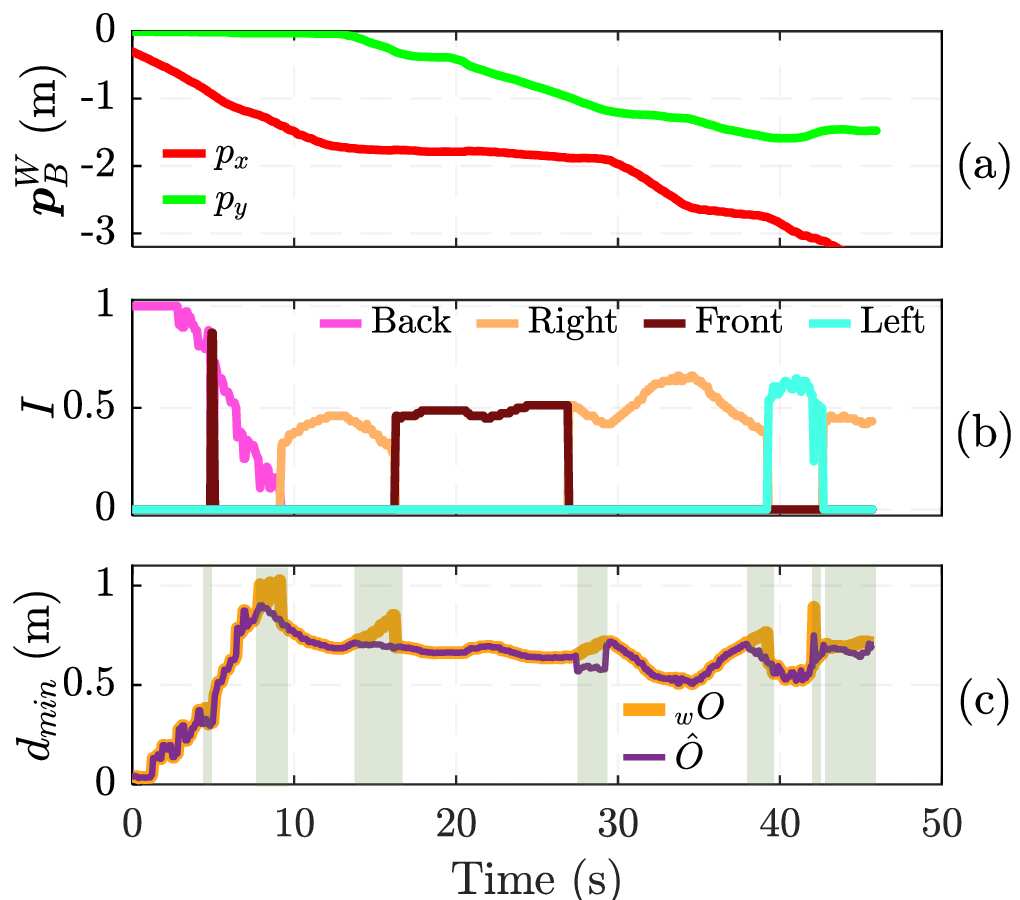}
	\caption{Results of the experiment where participant B co-transports the object when obstacle 2 is on the right. The graphs show (a) the position of the mobile base, (b) the intensity and the region of vibrotactile feedback provided to the human, and (c) the minimum distances to the obstacle to be warned (${}_{w}O$) and the closest obstacle ($\hat{O}$) where the green background indicates the cases when the Alg.~\ref{alg:cap} prevents switching between obstacles.}
    \vspace{-0.7cm}
	\label{fig:ergotac}
\end{figure}

In the second scenario, where this time obstacle 2 was located on the left side of the human, both participants continued to exhibit the same obstacle avoidance behaviors as in the first scenario. This means that participant A continued the initial motion without deviation until the front region of the belt vibrated intensely, while participant B preferred to switch the motion direction after feeling the vibrations indicating the presence of an obstacle on the left. Afterward, both participants reached the finish line without a significant change in their motion, even though participant B received slight and infrequent warnings from the front side.
These warnings, possibly resulting from noise in the LIDAR scan, did not prevent the user from moving toward the target due to their low intensity and short duration.

Moreover, when the vibration intensities are examined for both scenarios (the thickness of the colored lines), we see that participant B exhibits a more cautious attitude in approaching the obstacles with respect to participant A. Although this behavior prolongs the distance traveled to reach the target destination, the user opts for a longer path favoring safety.

Fig.~\ref{fig:ergotac} depicts the position of the robot base, generated vibration feedback, and the minimum distance of the ${}_{w}O$ (obstacle to be warned) and $\hat{O}$ (closest obstacle) to the footprint for the trial of participant B in the first scenario. The highlighted green background in Fig.~\ref{fig:ergotac}c indicates the intervals when Alg.1 prevents switching between obstacles to avoid unnecessary fluctuations between them.
When the vibration feedback and the highlighted areas are examined together, we see that even if a closer object in a different region was detected (see Alg.~\ref{alg:cap}), the obstacle warning module continued to inform the user about the current location of the previously warned object.  
Then, our proposed module only switched the region to be warned if the closest obstacle in the new region had a shorter $d_{min}$ than the previously warned one by a specified ratio.

\vspace{-0.1 cm}

\section{CONCLUSION}
\label{sec:conclusion}

In this work, we presented a novel haptic-based environmental awareness module for an HRC system for transporting large and heavy objects. The module has been integrated into our HRC transportation framework, expanding its features. The outcomes of this paper show how this new module enhances the situational awareness of the human operator acting as the leader by warning them of surrounding obstacles through a vibrotactile haptic belt. This belt was programmed to give information about the location and proximity of the obstacles measured by the robot's sensory system. The experimental evaluation results conducted with non-expert subjects indicated that the haptic feedback provided by the module allowed the human partner to carry out a co-transportation task. Hence, it was demonstrated how the proposed framework improved the safety of the human-robot team in co-carrying scenarios and enabled the human operator to lead the co-transportation system in an unknown environment with hidden obstacles. Future works will focus on the improvement of the autonomy of the robot partner to exploit the obstacle avoidance capabilities in an enhanced shared-autonomy human-robot framework.

\vspace{-0.15 cm}
\bibliographystyle{IEEEtran}
\bibliography{biblio.bib}
\end{document}